\documentclass[letterpaper]{article} 
\usepackage{aaai25}  
\usepackage{times}  
\usepackage{helvet}  
\usepackage{courier}  
\usepackage[hyphens]{url}  
\usepackage{graphicx} 
\usepackage{natbib}  
\usepackage{caption} 
\usepackage{commands} 
\usepackage{amsmath} 
\usepackage{amsthm} 
\usepackage{amssymb} 
\usepackage{latexsym} 
\usepackage{paralist} 
\usepackage{makecell} 
\usepackage{todonotes} 
\usepackage{algorithm}
\usepackage{algorithmic}

\usepackage{newfloat}
\usepackage{listings}
\DeclareCaptionStyle{ruled}{labelfont=normalfont,labelsep=colon,strut=off} 
\floatstyle{ruled}
\newfloat{listing}{tb}{lst}{}
\floatname{listing}{Listing}

\usepackage{xspace}

\newcommand{\myi}{(\emph{i})\xspace}
\newcommand{\myii}{(\emph{ii})\xspace}
\newcommand{\myiii}{(\emph{iii})\xspace}

\newcommand{\LTLf}{{\sc ltl}$_f$\xspace}

\newcommand{\LTL}{{\sc ltl}\xspace}

\newcommand{\limp}{\supset}
\newcommand{\lneg}{\neg}

\newcommand{\true}{\mathit{true}}

\newcommand{\false}{\mathit{false}}

\newcommand{\Next}{\raisebox{-0.27ex}{\LARGE$\circ$}}
\newcommand{\Wnext}{\raisebox{-0.27ex}{\LARGE$\bullet$}}
\newcommand{\Until}{\mathop{\U}}

\newcommand{\A}{\mathcal{A}}
\newcommand{\E}{\mathcal{E}}
\newcommand{\X}{\mathcal{X}}
\newcommand{\Y}{\mathcal{Y}}

\newcommand{\U}{\mathcal{U}}

\newcommand{\G}{\mathcal{G}}

\renewcommand{\L}{\mathcal{L}}
\renewcommand{\H}{\mathcal{H}}

\newcommand{\N}{\mathcal{N}}

\newcommand{\DFA}{{\sc dfa}\xspace}
\newcommand{\NFA}{{\sc nfa}\xspace}
\newcommand{\NFAs}{{\sc nfa}s\xspace}
\newcommand{\DFAs}{{\sc dfa}s\xspace}

\newcommand{\pspace}{{\sc pspace}\xspace}

\newcommand{\twoexptime}{2{\sc exptime}\xspace}

\newcommand{\trace}{\pi}

\newcommand{\ls}{{\sf{lst}}}
\newcommand{\tiff}{\text{ iff }}

\renewcommand{\stop}{\mathtt{stop}\xspace}
\newcommand{\naturals}{\mathbb{N}\xspace}

\newcommand{\tst}{\text{ s.t. }}

\newcommand{\existsweak}{\textsc{ExistsWeak}\space}
\newcommand{\checkweak}{\textsc{CheckWeak}\xspace}
\newtheorem{definition}{Definition}
\newtheorem{example}{Example}
\newtheorem{theorem}{Theorem}

\newcommand{\putaway}[1]{}
\usepackage{color}
\usepackage{enumitem}

\newcommand{\elnote}[1]{\footnote{EL: #1}}

\title{Computational Grounding of Responsibility Attribution and Anticipation in \LTLf}
\author {
    Giuseppe De Giacomo\textsuperscript{\rm 1,3},
    Emiliano Lorini\textsuperscript{\rm 2},
    Timothy Parker\textsuperscript{\rm 2}
    Gianmarco Parretti\textsuperscript{\rm 3}
}
\affiliations {
    \textsuperscript{\rm 1}University of Oxford, UK\\
    \textsuperscript{\rm 2}IRIT, CNRS, University of Toulouse, France\\
    University of Rome ``La Sapienza'', Italy\\
    giuseppe.degiacomo@cs.ox.ac.uk,
\{emiliano.lorini,timothy.parker\}@irit.fr,
parretti@diag.uniroma1.it}

\begin{document}

\maketitle

\begin{abstract}
Responsibility is one of the key notions in machine ethics
and in the area of autonomous systems. It is a multi-faceted notion
involving counterfactual reasoning about actions and strategies. 
In this paper,
we study  different variants
of responsibility 
 in a strategic setting based on \LTLf. 
 We show a connection with  notions in  reactive synthesis, including synthesis of winning, dominant, and best-effort strategies. 
This connection provides the building blocks for a computational grounding of responsibility including   
complexity characterizations and sound, complete, and optimal algorithms for attributing and anticipating responsibility.
\end{abstract}

%

\section{Introduction}

 Responsibility is a key notion in the areas of machine ethics and multi-agent systems. In recent times, there have been several and diverse attempts to formalize it, using various approaches such as game-theoretic tools \cite{Baier0M21,Braham2012,DBLP:journals/fuin/LoriniM18} and logical tools including STIT logic \cite{DBLP:journals/logcom/LoriniLM14,DBLP:conf/atal/AbarcaB22,Baltag2021-BALCAA-8,DBLP:journals/ai/LoriniS11}, \LTLf \cite{DBLP:conf/ecai/ParkerGL23},  ATL \cite{DBLP:conf/atal/YazdanpanahDJAL19,DBLP:conf/clima/BullingD13}, logics of strategic and extensive games \cite{DBLP:conf/aaai/Shi24,DBLP:conf/ijcai/Naumov021,DBLP:journals/apal/NaumovT23},
 structural equation models  \cite{DBLP:journals/jair/ChocklerH04}. However, the overall picture on the conceptual
 and computational  aspects of responsibility remains rather fragmented. This is due to its  polysemic nature
and to its  many dimensions  (e.g., forward-looking vs backward-looking, active vs passive, direct vs indirect, causal vs moral, attributed vs anticipated). 

In this paper we provide \myi a comprehensive analysis of the complexity of reasoning about strategic responsibility, namely, the responsibility of an agent due to its choice of a given strategy, and \myii  a number of algorithms that can be used to automate reasoning about strategic responsibility. Any ethical artificial agent should be endowed with this kind of reasoning.
First of all,
the agent should be able to \emph{attribute} responsibility to itself and to other agents in order to put in place reparatory actions, e.g.,  in case of responsibility for a norm violation or for a damage. Secondly, it should  be able to \emph{anticipate} its own responsibility and the responsibility of others in order to refrain from exposing  itself and others to possible blame and sanctions. 

To illustrate the relevance  of reasoning about strategic responsibility in ethical  AI applications, as well as to aid in the explanation of our model, we will use the running example of a robot who must take care of a plant. In the morning and afternoon, the robot has to decide whether or not to water the plant, knowing that it may also rain during that time. For simplicity, we will analyse only a single day.

Suppose that both
the robot and the environment decide not to water the plant under any condition, meaning that the plant will die since it was not watered. 
However, is the robot causally responsible for the death of the plant? Given the environment's choice, the robot could have prevented the
plant from dying by watering it, so 
the robot is responsible 
for the death of the plant
at least in a minimal sense.

The counterfactual requirement that one could have prevented 
$\omega$ by choosing differently
is called \emph{passive responsibility}  \cite{DBLP:journals/logcom/LoriniLM14}. It is put in opposition to 
\emph{active responsibility},
also called deliberative ``seeing to it that'' in the STIT tradition \cite{belnap01facing},  which consists
in an agent's choice forcing a certain
fact to be true regardless of the choices of the other agents. 
Passive responsibility is a notion of responsibility
in a weak sense, 
as the previous example highlights. 
As pointed out in \cite{Braham2012}, 
to strengthen it
one should add the requirement that  the alternative
 option the agent could have chosen
 was recommended or
 admissible 
 by some rationality requirement, 
 so that it has no excuse
 for what it chose. Two simple and universally accepted  rationality
  requirements are dominance and the related  notion of best-effort:
 there was an alternative recommended  
  option  in the sense that 
  the agent had 
an  alternative
choice that 
  dominates its  actual
  choice, which is equivalent to saying 
  that the agent's  actual
  choice is not best-effort. It comes as a welcome surprise 
  that such notions of dominance and best-effort
  have been independently developed in the field of strategy synthesis. So, it becomes natural
  to combine  
  the theory
  of responsibility
 with existing  work 
  on best-effort  synthesis.
  This is the focus of the present paper. 
What we gain by doing this  is a better understanding
of the complexity
of reasoning about strategic
responsibility and  novel  
algorithms for it.

In this paper we consider specifications expressed in Linear Time Logic on finite traces (\LTLf) \cite{DegVa13}. Focusing on these specifications, we give formal computational grounding to various forms of responsibility anticipation and attribution introduced in terms of \LTLf synthesis 
with a special
emphasis on dominance and best-effort,
given their relevance to responsibility.  Specifically, for each of the notions of responsibility attribution and  anticipation in this paper, we devise algorithms, inspired by \LTLf strategy synthesis and checking techniques, that are sound and complete, and characterize worst-case computational complexity of the problem. This array of results provides novel computational foundations for a comprehensive responsibility analysis
on temporal specifications.

\section{Preliminaries~\label{sec:preliminaries}}


A \emph{trace} over 
an alphabet of symbols 
$\Sigma$ is a finite or infinite
sequence of elements from $\Sigma$. The empty trace is
$\lambda$. The length of a trace is $|\trace|$.
Traces are indexed starting at zero, and we write $\trace = \trace_0 \trace_1 \cdots$. For a finite trace $\trace$, we denote by $\ls(\trace)$ the index of its last element, i.e., $|\trace| - 1$. 
We denote by $\trace^k = \trace_0 \cdots \trace_k$ the prefix of $\trace$ up to the $k$-th index.


\textit{Linear Temporal Logic on finite traces}~(\LTLf) is a specification language for expressing temporal properties over finite traces~\cite{DegVa13}. \LTLf has the same syntax as \LTL, which is instead interpreted over infinite traces~\cite{Pnu77}. Given a set $AP$ of atomic propositions (aka atoms), the \LTLf formulas over $AP$ are: \\ 
\centerline{$\omega ::= a \mid \neg \omega \mid \omega \wedge \omega \mid 
 \Next \omega \mid \omega \Until \omega$}
Where $a \in AP$, and $\Next$~(\emph{Next}) and $\Until$~(\emph{Until}) are temporal operators. 
We use standard Boolean abbreviations $\vee$~(or) $\supset$~(implies), $\true$, and $\false$.
We also use the following abbreviations: $\Wnext \omega \equiv \neg \Next \neg \omega$ (\emph{Weak Next}); $\Diamond \omega \equiv \true \Until \omega$ (\emph{Eventually}); and $\Box \omega \equiv \neg \Diamond \neg \omega$ (\emph{Always}).
The size of $\omega$, written $|\omega|$, is the number of its subformulas.

\LTLf formulas are interpreted over finite traces $\pi$ over the alphabet $\Sigma = 2^{AP}$, i.e., consisting of propositional interpretations of atoms. For $i \leq \ls({\pi})$, we have that $\trace_i \in 2^{AP}$ is the $i$-th interpretation of $\trace$. That an \LTLf formula $\omega$ \emph{holds} at instant $i \leq \ls(\pi)$, written $\trace, i \models \omega$, is defined inductively: 1. $\trace, i \models a \tiff a \in \trace_i\nonumber$ (for $a\in AP$); 2. $\trace, i \models \lnot \omega \tiff \trace, i \not\models \omega\nonumber$; 3. $\trace, i \models \omega_1 \wedge \omega_2 \tiff \trace, i \models \omega_1 \text{ and } \trace, i \models \omega_2\nonumber$; 4. $\pi, i \models \Next\omega \tiff  i< \ls(\pi)$ and $\trace,i+1 \models \omega$; and 5. $\pi, i \models \omega_1 \Until \omega_2$ iff $\exists j$ such that $i \leq j \leq \ls(\pi)$ and $\pi,j \models\omega_2$, and $\forall k, i\le k < j$ we have that $\pi, k \models \omega_1$. We say that $\pi$ \emph{satisfies} $\omega$, written $\pi \models \omega$, if $\pi, 0 \models \omega$.

\LTLf \emph{reactive synthesis under environment specifications} \cite{DegVa15,AminofDMR19} concerns finding a strategy to satisfy an \LTLf formula while interacting with an environment.
We consider
\LTLf formulas over $AP = \Y \cup \X$, where $\Y$ and $\X$ are disjoint sets of atoms under control of agent and environment, respectively. Traces over $\Sigma = 2^{\Y \cup \X}$ will be denoted $\pi = (Y_0 \cup X_0)(Y_1 \cup X_1)\cdots$ where $Y_i \subseteq \Y$ and $X_i \subseteq \X$ for every $i \geq 0$. 


An \emph{agent strategy} is a function $\sigma_{ag}: (2^{\X})^* \rightarrow 2^{\Y}$ mapping sequences of environment choices to an agent choice. We require $\sigma_{ag}$ to be \emph{stopping}~\cite{DeGiacomoDPZ21}, i.e., the agent stops the execution of any action at some point of the trace, 
written $\stop$.
Formally, an agent strategy $\sigma_{ag}$ is stopping if for every trace $\trace \in (2^{\Y \cup \X})^\omega$ there exists $k \in \naturals$ such that, for every $j \geq k$, we have that $\sigma_{ag}(\trace^j) = \stop$ 
and, for every $i < k$, we have that $\sigma_{ag}(\trace^i) \neq \stop$.
An \emph{environment strategy} is a function $\sigma_{env}: (2^\Y)^+ \rightarrow 2^{\X}$ mapping non-empty sequences of agent choices to an environment choice. The domain of $\sigma_{ag}$ includes the empty sequence $\lambda$ as we assume that the agent moves first.

A trace $\trace = (Y_0 \cup X_0) \cdots (Y_n \cup X_n)$ is consistent with $\sigma_{ag}$ if: \myi $Y_0 = \sigma(\lambda)$; \myii $Y_i = \sigma_{ag}(X_0 \cdots X_{i-1})$ for every $i > 0$; and \myiii $\sigma_{ag}(X_0 \cdots X_n) = \stop$. Similarly, $\trace$ is consistent with $\sigma_{env}$ if $X_j = \sigma_{env}(Y_0 \cdots Y_j)$ for every $j \geq 0$. We denote by $\play(\sigma_{ag}, \sigma_{env})$ the shortest trace that is consistent with both $\sigma_{ag}$ and $\sigma_{env}$, called \emph{play}.


Let $\psi$ be an \LTLf formula over $\Y \cup \X$. An agent strategy $\sigma_{ag}$ is \emph{winning} for (aka \emph{enforces}) $\psi$ if $\play(\sigma_{ag}, \sigma_{env}) \models \psi$ for every environment strategy $\sigma_{env}$. Conversely, an environment strategy $\sigma_{env}$ is \emph{winning} (aka \emph{enforces}) for $\psi$ if \emph{every} finite prefix of $\play(\sigma_{ag}, \sigma_{env})$ satisfies $\psi$ for every agent strategy $\sigma_{ag}$. An environment specification $\E$ is an \LTLf formula that is environment enforceable. We denote by $\Sigma_{\E}$ the set of environment strategies that enforce $\E$. 

Let $\omega$ be an \LTLf formula and $\E$ an \LTLf environment specification. Reactive synthesis under environment specifications is the problem of finding an agent strategy $\sigma_{ag}$ such that $\play(\sigma_{ag}, \sigma_{env})\models\omega$ for every environment strategy $\sigma_{env} \in \Sigma_{\E}$, if one exists~\cite{AminofDMR19}. Such a strategy is called winning for $\omega$ under $\E$.

We also consider \emph{dominant}  and \emph{best-effort} strategies~\cite{AminofDR21,AminofDR2023}, which are based on the game-theoretic notion of dominance~\cite{AptG2011}. Let $\sigma_1$ and $\sigma_2$ be agent strategies. We say that $\sigma_1$ \emph{dominates} $\sigma_2$ for $\omega$ under $\E$, written \mbox{$\sigma_1 \geq_{\omega|\E} \sigma_2$}, if for every environment strategy $\sigma_{env} \in \Sigma_{env}$, if $\play(\sigma_2, \sigma_{env}) \models \omega$, then $\play(\sigma_1, \sigma_{env}) \models \omega$. Furthermore, we say that $\sigma_1$ \emph{strictly dominates} $\sigma_2$, written \mbox{$\sigma_1>_{\omega|\E} \sigma_2$}, if $\sigma_1 \geq_{\omega|\E} \sigma_2$ and $\sigma_2 \not \geq_{\omega|\E} \sigma_1$. Intuitively, $\sigma_1 >_{\omega|\E} \sigma_2$ means that $\sigma_1$ is at least as good as $\sigma_2$ against every environment strategy enforcing $\E$ and strictly better against at least one such strategy.


%
An agent strategy $\sigma_{ag}$ is \emph{dominant} for $\omega$ under $\E$ if for every agent strategy $\sigma'_{ag}$  we have that $\sigma_{ag} \geq_{\omega|\E} \sigma'_{ag}$. 

An agent strategy $\sigma_{ag}$ is best-effort for $\omega$ under $\E$ if there does not exist an agent strategy $\sigma'_{ag}$ such that $\sigma'_{ag} >_{\omega|\E} \sigma_{ag}$. 
Unlike winning and dominant strategies, best-effort strategies always exist~\cite{AminofDR21}. 


Winning, dominant, and best-effort strategies are related as follows \cite{AminofDR2023}: every winning strategy is dominant and every dominant strategy is best-effort; if there exists a winning strategy, dominant strategies are winning; if there exists a dominant strategy, best-effort strategies are dominant.
Best-effort and dominant strategies also admit a
\emph{local characterization} that uses the notion of \emph{value} of a history~\cite{AminofDR21,AminofDR2023}, i.e., finite sequence of agent and environment moves.
Intuitively, the value of a history $h$ is: $+1$ (``winning"), if the agent can enforce $\omega$ under $\E$ from $h$; $-1$ (``losing") if the agent can never satisfy $\omega$ under $\E$ from $h$; and $0$ (``pending") otherwise. With this notion, best-effort
strategies witness the maximum value of each
history $h$ consistent with them. Dominant strategies witness that, for pending histories $h$ (that do not extend winning ones), exactly one agent move $Y$ allows extending $h$ so that the extended history $h \cdot Y$ is not losing.

Let $h$ be a history and $\strategyag$ an agent strategy. We denote by $\Sigma_{\E}(h, \sigma_{ag})$ the set of environment strategies $\sigma_{env}$ enforcing $\E$ such that $h$ is consistent with $\sigma_{ag}$ and $\sigma_{env}$. Also, we denote by $\H_{\E}(\sigma_{ag})$ the set of all histories $h$ such that $\Sigma_{\E}(h, \sigma_{ag})$ is non-empty, i.e., $\H_{\E}(\sigma_{ag})$ is the set of all histories that are consistent with $\sigma_{ag}$ and some environment strategy enforcing $\E$. For $h \in \H_{\E}(\sigma_{ag})$, define:
\begin{compactenum}
    \item $val_{\omega|\E}(\sigma_{ag}, h) = +1$ (``winning") if $\play(\sigma_{ag}, \sigma_{env})$ is finite and satisfies $\omega$ for every $\sigma_{env} \in \Sigma_{\E}(h, \sigma_{ag})$;
    \item $val_{\omega|\E}(\sigma_{ag}, h) = -1$ (``losing") if $\play(\sigma_{ag}, \sigma_{env})$ is finite and satisfies $\lneg \omega$ for every $\sigma_{env} \in \Sigma_{\E}(h, \sigma_{ag})$;
    \item $val_{\omega|\E}(\sigma_{ag}, h) = 0$ (``pending").
\end{compactenum} 
We denote by $val_{\omega|\E}(h)$ the maximum of $val_{\omega|\E}(\sigma_{ag}, h)$ over all $\sigma_{ag}$ such that $h \in \H_{\E}(\sigma_{ag})$ (we define $val_{\omega|\E}(h)$ only in case $h \in \H_{\E}(\sigma_{ag})$ for some $\sigma_{ag}$). Here is the local characterization of best-effort and dominant strategies~\cite{AminofDR2023}: 
\begin{compactitem}
    \item An agent strategy $\sigma_{ag}$ is best-effort for $\omega$ under $\E$ iff $val_{\omega|\E}(\sigma_{ag}, h) = val_{\omega|\E}(h)$ for every $h \in \H_{\E}(\sigma_{ag})$;
    \item Furthermore, $\sigma_{ag}$ is dominant for $\omega$ under $\E$ iff $val_{\omega|\E}(h) = 0$ implies $val_{\omega|\E}(h \cdot Y) = -1$ for every $h \in \H_{\E}(\sigma_{ag})$ ending in an environment move (including the empty history $\lambda$)
    and $Y \neq \sigma_{ag}(h)$.
\end{compactitem}

\emph{Dominant} and \emph{best-effort} synthesis under environment specifications are the problems of finding a dominant and best-effort strategy for $\omega$ under $\E$, if one exists, respectively.  
    Reactive, dominant, and best-effort synthesis are \twoexptime-complete \cite{DegVa15,AminofDMR19,AminofDR21,AminofDR2023}.

We will also need to consider \emph{weak} strategies, a form of strategy taken from planning in nondeterministic domains~\cite{CimattiPRT03} that captures the possibility for the agent to satisfy an \LTLf formula. Formally, an agent strategy $\sigma_{ag}$ is \emph{weak} for $\omega$ under $\E$ if there exists an environment strategy $\sigma_{env} \in \Sigma_{\E}$ such that $\play(\sigma_{ag}, \sigma_{env}) \models \omega$.

\section{Responsibility}\label{sec:responsibility}

The paper studies the notion of causal responsibility \cite{Vincent2011}, which specifies  whether or not an agent 
caused  a certain state of affairs
to occur by making a certain choice. It is a general notion of responsibility that is a necessary condition for both legal \cite{Jansen14} and moral \cite{sep-moral-responsibility} responsibility. We will focus on two forms of responsibility that were considered in the literature, active responsibility and passive responsibility. Active responsibility captures the notion of an agent making $\omega$  happen, while passive responsibility consists in the agent merely letting $\omega$ happen.
The distinction between the two notions was proposed  in 
\cite{DBLP:journals/logcom/LoriniLM14} in an action-based setting. 
Active responsibility corresponds to the notion of \emph{deliberative stit} studied in STIT logic while passive responsibility
corresponds to the counterfactual  notion of \emph{(something) could have been prevented} (CHP), a fundamental component of the notion of regret 
as highlighted in \cite{DBLP:journals/ai/LoriniS11}. More recently, a plan-based analysis of active and passive responsibility was proposed in 
\cite{DBLP:conf/ecai/ParkerGL23}. In line with their work
we distinguish attribution from
 anticipation
for both active and passive responsibility.
Responsibility attribution is an \emph{ex post}
notion: it is ascribed to a given agent after the agent and the environment  have made their choices and the result of their choices has been revealed. 
Responsibility anticipation is an \emph{ex ante}
notion: it is the responsibility that an agent \textit{could} incur by making a certain choice.

In this section, we present a novel
strategy-based analysis of 
active and passive responsibility,
from the point of view of both attribution and anticipation. When moving from actions and linear plans to strategies some conceptual issues arise. In particular, when attributing   passive responsibility
for a fact $\omega $ to an agent, one has to fix the choice
made by the environment and counterfactually check whether 
the agent could have prevented $\omega$ from being true
by making a different choice.
While
the action or plan chosen by the environment can be easily extracted from the actual history which we assume to be fully observable, 
the same cannot be done for strategies. 
From the actual history, one can only infer
the set of strategies
of the environment that are consistent with it
and this set 
is not necessarily a singleton. 
So, in a strategy-based setting we must distinguish responsibility
attribution against agent strategies (supposing the environment strategy is fully observable) from 
responsibility
attribution
on histories
(when the environment strategy is not fully known).

  

Another novel aspect  of our analysis, in addition to the focus on strategies and the consequent distinction between  responsibility attribution against environment strategies or on histories, is a refinement of the notion  of passive responsibility. 
Passive responsibility, as defined in
\cite{DBLP:journals/logcom/LoriniLM14,DBLP:conf/ecai/ParkerGL23}, 
requires
to existentially quantify over
the agent's  possible  choices. 
We consider a variant of passive responsibility attribution and anticipation
in which existential
quantification is restricted to the set of possible  choices
that dominates the agent's actual choice with respect to the avoidance of $\omega$. 
We call this \emph{inexcusable} passive responsibility. 
 Specifically,  an agent is held responsible for letting $\omega$
be true \emph{with no excuse}, if 
it could have prevented $\omega$ from being true
by  choosing a different strategy 
that dominates its  actual choice with respect
to the avoidance of $\omega$. In other words (as we will prove later),
the agent chose a strategy which is not best-effort for the avoidance of $\omega $. 
We say that the agent has no excuse
since  it cannot appeal to not knowing the choice of the environment
at the moment of its choice, since it had an alternative choice available 
that would have guaranteed the avoidance
of $\omega $
\emph{more than} the choice it actually made,
irrespective of the choice of the environment.
 
In what follows we denote \LTLf temporal properties as $\omega$, \LTLf environment specifications as $\E$, agent strategies as $\sigma_{ag}$, environment strategies as $\sigma_{env}$, and histories as $h$.







\paragraph{Attribution against Environment Strategies.}
We first define responsibility attribution in a perfect information setting with full knowledge of the environment strategy. While convenient, this is uncommon in a strategic setting where the environment is the model of the world in which the agent operates. Nonetheless, these notions are useful for defining further refinements of responsibility. 


\begin{definition}[Passive Responsibility Attribution, \textsc{PRAttr}]\label{def:pass-resp-attr}
The agent is attributed passive responsibility for $\omega$ under $\strategyag$ and $\strategyenv \in \Sigma_\E$ if $\play(\strategyag, \strategyenv) \models \omega$ and there exists an agent strategy $\strategyag'$ such that $\play(\strategyag', \strategyenv) \models \neg \omega$.
\end{definition}

\begin{definition}[Inexcusable Passive Responsibility Attribution, \textsc{IPRAttr}]\label{def:inex-pass-resp-attr}
The agent is attributed inexcusable passive responsibility 
for $\omega$ under $\strategyag$ and $\strategyenv \in \Sigma_\E$ if $\play(\strategyag, \strategyenv) \models \omega$ and there exists an agent strategy $\strategyag'$ such that $\strategyag' \geq_{\lneg \omega|\E} \strategyag$ and $\play(\strategyag', \strategyenv) \models \neg  \omega$.
\end{definition}

We assume 
 that at the moment of its choice
 the agent 
 has no information about the possible
 choices of the environment beyond the environment specification.
This is because in our model the environment is not necessarily rational
and the agent knows this.
Consequently, according to the agent, 
all environment strategies
enforcing the environment specification 
are possible, regardless of apparent rationality. 
This also justifies the use of (weak) dominance 
and best-effort for characterizing 
the notion
of `inexcusable' in the previous definition. Indeed, 
if there exists
$\strategyag'$
for the agent
which dominates the strategy 
$\strategyag$
with respect to  $\neg \omega$,
the choice of $\strategyag$ is inexcusable since
it  has certainly envisaged an
environment strategy 
for which the alternative strategy $\strategyag'$ would have been strictly better than the strategy $\strategyag$.


\paragraph{Attribution on Histories.} We now consider the imperfect information setting where we have access only to the history, which is far less problematic than 
access to the full environment strategy. In the following definitions, we assume that the history $h$ is consistent with the strategy $\sigma_{ag}$.

\begin{definition}[(Inexcusable) Passive Responsibility Attribution on History, \textsc{IPRAttr$(h)$/PRAttr}$(h)$]
The agent is attributed (inexcusable) passive responsibility for $\omega$ under $\sigma_{ag}$, $\E$, and $h$ if there is an environment strategy $\sigma_{env} \in \Sigma_{\E}$ s.t $h$ is consistent with $\sigma_{env}$ and the agent is attributed (inexcusable) passive responsibility  
for $\omega$ under $\sigma_{ag}$ and $\sigma_{env}$.
\end{definition}






\paragraph{}

Responsibility anticipation is ex ante, meaning that it reasons in terms of potential responsibility that an agent could incur by making a certain choice. This means that is quantifies over the set of possible environment strategies rather than any single environment strategy, meaning it only requires access to $\E$. Therefore there is no difference between responsibility anticipation in a perfect information setting and an imperfect information setting.

\begin{definition}[(Inexcusable) Passive Responsibility Anticipation, \textsc{IPRAnt/PRAnt}]
The agent anticipates (inexcusable) passive responsibility for $\omega$ under $\strategyag$ and $\E$ if there is an environment strategy $\strategyenv \in \Sigma_\E$ s.t. the agent is attributed (inexcusable) passive responsibility for $\omega$ under $\strategyag$ and $\strategyenv$.
\end{definition}


The agent anticipates responsibility for $\omega$ iff there is some possible outcome where it is attributed responsibility for $\omega$. It follows that if the agent adopts a strategy that does not anticipate X-responsibility then the strategy cannot be attributed X-responsibility against any environment strategies or on any history (for any form of responsibility X).

\paragraph{Active Responsibility}
Since active responsibility quantifies over environment strategies, the processes for attributing responsibility (against histories or strategies) and anticipating responsibility are all one and the same. Correspondingly we give a single definition for active responsibility. 

\begin{definition}[Active Responsibility Attribution and Anticipation, \textsc{ARA}]\label{def:act-resp}
The agent is attributed (anticipates) active responsibility for $\omega$ under $\strategyag$ and $\E$ if \mbox{$\play(\strategyag, \strategyenv) \models \omega$} for every environment strategy $\strategyenv \in \Sigma_{\E}$ and there exists a pair of strategies $(\strategyag',\strategyenv')$ such that $\strategyenv' \in \Sigma_{\E}$ and $\play(\strategyag', \strategyenv') \models \neg \omega$.
\end{definition}

The second condition ensures that the agent is not actively responsible for an inevitable outcome (e.g., the sunrise). 

\section{Responsibility and Strategy Properties}


Based on their definitions, active, passive and inexcusable passive responsibility are strongly connected with winning, dominant and best-effort strategies, respectively.

\begin{theorem}\label{thm:pass-and-dom}~
\begin{compactitem}
        \item The agent anticipates passive (resp. inexcusable passive) responsibility for $\omega$ under $\strategyag$ and $\E$ iff $\strategyag$ is not dominant (resp. is not best-effort) for $\lnot \omega$ under $\E$.
        \item The agent anticipates active responsibility for $\omega$ under $\strategyag$ and $\E$ iff $\strategyag$ is winning for $\omega$ under $\E$ and there exists some $\strategyag'$ that is weak for $\lneg \omega$ under $\E$;
    \end{compactitem}
\end{theorem}

By Theorem~\ref{thm:pass-and-dom} and the relation between best-effort, dominant, and winning strategies, an agent using a best-effort strategy $\sigma_{ag}$ for some goal or value $\omega$ will not anticipate inexcusable passive responsibility for $\neg \omega$. The agent will also not anticipate passive responsibility for $\neg \omega$ if possible (i.e. if any strategy does not anticipate passive responsibility)
and will anticipate active responsibility for $\omega$ if possible (i.e. if any strategy anticipates active responsibility). 
For this reason, and because best-effort strategies always exist, we argue that inexcusable passive responsibility is the most useful  
notion of causal responsibility: an agent that only considers inexcusable passive responsibility for strategy evaluation will never need to consider passive or active responsibility.

There also exists a similar connection between passive and inexcusable passive responsibility attribution on histories and dominant and best-effort strategies, respectively. With responsibility attribution on a history $h$, environment strategies are restricted to those such that $h$ is consistent with them. We show that we can capture exactly the set of such strategies with a suitable \LTLf environment specification.

Let $h = (Y_0 \cup X_0) \cdots (Y_n \cup X_n)$ be a history. That $h$ is consistent with an environment strategy $\strategyenv$ means that $X_i = \strategyenv(Y_0 \cdots Y_{i})$ for every $i \geq 0$. Such behavior can be captured by the following \LTLf environment specification. 
$$
\begin{array}{l}
\E_h = (Y_0 \limp X_0) \land ((Y_0 \land \Wnext^1 Y_1) \limp \Wnext^1 X_1)~\land {}\\
\cdots~\land~((Y_0 \land \Wnext^1 Y_1 \land \cdots \land \Wnext^n Y_n) \limp \Wnext^n X_n)
\end{array}
$$
where $\Wnext^k$ denotes $k$ nested $\Wnext$ operators. It is easy to see that $h$ is consistent with $\sigma_{env}$ iff $\sigma_{env} \in \Sigma_{\E_h}$. With this result, we can  prove the following:

\begin{theorem}\label{thm:pass-and-dom-2}
    The agent is attributed passive (resp. inexcusable passive) responsibility for $\omega$ under $\strategyag$, $\E$, and $h$ iff $\strategyag$ is not dominant (resp. best-effort) for $\lneg \omega$ under $\E \land \E_h$.
\end{theorem}


By exploiting  Theorems~\ref{thm:pass-and-dom} and~\ref{thm:pass-and-dom-2}, we can give the computational grounding to the various responsibility notions (the only exceptions being \textsc{PRAttr} and \textsc{IPRAttr}, discussed later in the paper), whose computational complexity is shown in Table~\ref{table:responsibility-checking}. Membership of each responsibility notion is established from its corresponding algorithm, see Table~\ref{table:responsibility-checking}, whose correctness follows from Theorems~\ref{thm:pass-and-dom} and~\ref{thm:pass-and-dom-2}. Hardness follows by using Theorems~\ref{thm:pass-and-dom} and~\ref{thm:pass-and-dom-2} to reduce from suitable strategy checking problems, i.e., checking if a strategy is winning, dominant or best-effort, discussed in Checking Techniques, see Theorems~\ref{thm:check-win-complexity},~\ref{thm:check-dom-complexity}, and~\ref{thm:best-effort-checking-complexity}.



\begin{table}[t]
    \centering
    \resizebox{.99\linewidth}{!}
    {\begin{tabular}{|l||l|l|}
    \hline
        &\textbf{Complexity}$(\omega, \E, \sigma_{ag})$&\textbf{Algorithm$(\omega, \E, \sigma_{ag})$}\\
    \hline \hline
    \textsc{PRAnt} & \makecell[l]{\pspace-C $(\omega)$ \\
    \twoexptime-C $(\E)$\\ 
    poly $(\sigma_{ag})$} & \makecell[l]{$\lneg \checkdom(\lneg \omega, \E, \strategyag)$} \\ \hline
    \makecell[l]{\textsc{IPRAnt}} & \makecell[l]{\twoexptime-C [$\omega$][$\E$] \\ poly $(\sigma_{ag})$} & \makecell[l]{$\lneg \checkbe(\lneg \omega, \E, \sigma_{ag})$} \\ \hline
    \textsc{PRattr}$(h)$ & \makecell[l]{\pspace-C 
 $(\omega)$ \\
    \twoexptime-C $(\E)$\\ 
    poly $[\sigma_{ag}][h]$} & \makecell[l]{$\lneg \checkdom(\lneg \omega, \E \land \E_h, \sigma_{ag})$} \\ \hline
    \textsc{IPRattr}$(h)$ & \makecell[l]{
    \twoexptime-C $[\omega][\E]$\\ 
    poly $[\sigma_{ag}][h]$} & \makecell[l]{$\lneg \checkbe(\lneg \omega, \E \land \E_h, \sigma_{ag})$}\\ \hline
    \textsc{ARA} & \makecell[l]{\pspace-C 
 $(\omega)$\\
    \twoexptime-C $(\E)$\\ 
    poly $(\sigma_{ag})$} & \makecell[l]{$\checkwin(\omega, \E, \strategyag) \land{}$\\$\textsc{ExistsWeak}(\lneg \omega, \E)$} \\ \hline
    \end{tabular}}
    \caption{Computational grounding (complexity and algorithm) of passive and active responsibility anticipation and passive responsibility attribution on histories.}
    \label{table:responsibility-checking}
\end{table}

\begin{theorem}\label{thm:responsibility-checking-thm}
    The worst-case computational complexity of the various responsibility notions is established in Table~\ref{table:responsibility-checking}.
\end{theorem}

\noindent

\noindent \textbf{Example 1.} 
Before turning to the algorithms for checking responsibility, we
show how to use these notions in practice using the plant watering example from the introduction.

     Consider the following temporal properties: \begin{compactitem}
        \item $\varphi_1 $ = ``The plant has been watered at least once";
        \item $\varphi_2 $ = ``The plant has been watered exactly once";
        \item $\varphi_3$ = ``The plant has not been watered at all"
    \end{compactitem}
    The agent can decide to water the plant morning, evening, or both, and has the following strategies (amongst others): \begin{compactitem}
        \item $\sigma_1$ = ``Water the plant morning and evening";
        \item $\sigma_2$ = ``Water the plant only in the morning";
        \item $\sigma_3$ = ``Never water the plant";
    \end{compactitem}
    Consider the environment specifications and strategies: \begin{compactitem}
        \item $\E_1 =$``The weather can rain at any time of the day";
        \item $\strategyenv =$ ``The weather rains only in the evening";
    \end{compactitem}
    With $\sigma_1$ the agent is attributed (and  anticipates) active responsibility for $\varphi_1$ under $\E_1$, as $\sigma_1$ is winning for $\varphi_1$ under $\E_1$. With $\sigma_3$ the agent is not attributed active responsibility for $\varphi_1$ under $\E_1$, as $\sigma_{3}$ is not winning for $\varphi_1$ under $\E_1$, i.e., $\sigma_3$ is weak for $\lneg \varphi_1$ (the plant is never watered) under $\E$.
    

    With $\sigma_2$ the agent is attributed passive responsibility for $\lneg \varphi_2$, (either the plant has not been watered at all or has been watered more than once), against the environment strategy $\sigma_{env}$, since if the agent used $\sigma_3$ instead (fixing $\strategyenv$), then $\varphi_2$ would have been satisfied. 
    
    With $\sigma_{2}$, the agent does not anticipate (i.e., can \emph{not} be attributed) inexcusable passive responsibility for $\lneg \varphi_2$ and $\E_1$, as $\sigma_2$ is best-effort for $\varphi_2$ under $\E_1$ (but not dominant, as it does not dominate $\sigma_3$). That also applies for $\sigma_3$, as they are both best-effort for $\varphi_2$ under $\E_1$ (but not dominant).

    With $\sigma_3$ the agent does not anticipate passive responsibility (nor inexcusable passive responsibility) for $\lneg \varphi_3$, (the plant has been watered at least once) under $\E_1$, as $\sigma_3$ is dominant for $\varphi_3$ under $\E_1$ ($\sigma_3$ is the only strategy that satisfies $\varphi_3$ should the weather never rain and dominates $\sigma_1$ and $\sigma_2$). 

\section{Checking Techniques}\label{sec:checking}

The rest of the paper is devoted to presenting the algorithms for responsibility attribution and anticipation shown in Table~\ref{table:responsibility-checking}. The building blocks of these algorithms are finite-state automata and \DFA games, which we review below.   

A \emph{nondeterministic finite automaton} (\NFA) is a tuple $\N = (\Sigma, S, s_0, \delta, F)$, where: $\Sigma$ is a finite input alphabet; $S$ is a finite set of states; $s_0 \in S$ is the initial state; $\delta: S \times \Sigma \rightarrow 2^{S}$ is the transition function; and $F \subseteq S$ is the set of final states. The size of $\N$ is $|S|$. Given a word $\pi = \pi_0 \cdots \pi_n \in \Sigma^*$, a \emph{run} of $\N$ in $\pi$ is a sequence of states $s_0 \cdots s_{n+1}$ starting in the initial state of $\N$ and such that $s_{i+1} \in \delta(s_i, \pi_i)$ for every $i \geq 0$. A word $\pi$ is \emph{accepted} by $\N$ if it has a run whose last reached state is final. 
The language of $\N$, written $\L(\N)$, is the set of words accepted by $\N$. An automaton $\N$ is a \emph{deterministic finite automaton} (\DFA) if $|\delta(s, a)| \leq 1$ for every $(s, a) \in S \times \Sigma$. Checking non-emptiness of $\L(\N)$, written $\textsc{NonEmpty}(\N)$, can be done by checking the existence of a path from the initial state of $\N$ to some final state. Given the \NFAs $\N_1$ and $\N_2$ with languages $\L(\N_1)$ and $\L(\N_2)$, respectively, we can build in polynomial time the product \NFA $\N = \N_1 \times \N_2$ such that $\L(\N) = \L(\N_1) \cap \L(\N_2)$. We denote nondeterministic and deterministic automata by $\N$ and $\A$, respectively. Every \LTLf formula $\omega$ can be transformed into an \NFA $\N_{\omega}=\textsc{ToNFA}(\omega)$ (resp. \DFA $\A_{\omega}=\textsc{ToDFA}(\omega)$) with size at most exponential (resp. doubly-exponential) in $|\omega|$ and whose language is exactly the set of traces satisfying $\omega$~\cite{DegVa15}.


A \DFA game is a \DFA $\G$ with input alphabet $2^{\Y \cup \X}$, where $\Y$ and $\X$ are two disjoint sets under control of agent and environment, respectively. The notions of plays and strategies in Preliminaries also apply to \DFA games. An agent strategy is \emph{winning} if $\play(\sigma_{ag}, \sigma_{env})$ is accepted by $\G$ for every environment strategy $\sigma_{env}$. An agent strategy is \emph{weak} if $\play(\sigma_{ag}, \sigma_{env})$ is accepted by $\G$ for some environment strategy $\sigma_{env}$. Conversely, an environment strategy is \emph{winning} if $\play(\sigma_{ag}, \sigma_{env})$ is not accepted by $\G$ for every agent strategy 
$\sigma_{ag}$. The \emph{agent winning} (resp. \emph{weakly winning}) \emph{region} is the set of states $s \in S$ for which the agent has a winning (resp. weak) strategy in the game $\G' = (2^{\Y \cup \X}, S, s, \delta, F)$, i.e., the same game as $\G$, but with initial state $s$. The \emph{environment winning region} is defined analogously.  \emph{Solving} a \DFA game is the problem of computing the winning (resp. weakly winning) region, written $W = \textsc{WinRegion}(\G)$ (resp. $W' = \textsc{WeakRegion}(\G)$). Games played over \DFAs are determined, meaning that the agent winning region and the environment winning region partition the state space~\cite{gale1953infinite}. The environment winning region is denoted $\textsc{EnvWin}(\G)$. \DFA games can be solved in polynomial time in the size of the game via a least-fixpoint computation~\cite{AptG2011}. 

Sometimes we need to restrict transitions in a \DFA to those that do not allow leaving a given set of states. Given a \DFA $\A$ and a set of states $Q \subseteq S$, the restriction of $\A$ to $Q$ can be built in polynomial time and is denoted $\A' = \textsc{Restr}(\A, Q)$. We can reason about all environment strategies enforcing an \LTLf environment specification $\E$ by restricting its \DFA $\A_{\E}$ to the environment winning region $\textsc{EnvWin}(\A_\E)$.  

We represent agent strategies as \emph{terminating transducers}~\cite{BansalLTVW23} $\sigma_{ag} = (2^{\X}, 2^{\Y}, S, s_0, \eta, \kappa, F)$, where: $2^{\X}$ is the input alphabet; $2^{\Y}$ is the output alphabet; $S$ is a finite set of states; $s_0 \in S$ is the initial state; $\eta: S \times 2^{\X} \rightarrow S$ is the transition function; $\kappa: S \rightarrow 2^{\Y}$ is the output function; and $F \subseteq S$ is the set of terminating states. The size of $\sigma_{ag}$ is $|S|$. Given an input sequence $X_0 \cdots X_n \in (2^{\X})^*$, the output sequence is $\kappa(s_0) \cdots \kappa(s_n)$, where $s_0$ is the initial state of $\sigma_{ag}$ and $s_{i+1} = \eta(s_i, X_i)$ for every $i \geq 0$. 
A transducer $\sigma_{ag}$ can be transformed in polynomial time into a \DFA $\A_{\sigma_{ag}} = \textsc{ToDFA}(\sigma_{ag})$ whose language is the set of traces consistent with $\sigma_{ag}$.
We now give algorithms for checking that a strategy is of a specific type, i.e., winning, dominant, or best-effort, and the existence of weak strategies,
which form the techniques for responsibility attribution and anticipation shown in Table~\ref{table:responsibility-checking}.




We present below an algorithm to decide if a weak strategy for $\omega$ under $\E$ exists, denoted $\textsc{ExistsWeak}(\omega, \E)$: 1. Construct the \NFA $\N_{\omega}$ of $\omega$ and \DFA $\A_{\E}$ of $\E$; 2. restrict $\A_{\E}$ to the environment winning region and obtain \DFA $\A'_{\E}$; and 3. check language non-emptiness of the product \mbox{$\N = \N_{\omega} \times \A'_{\E}$}. That is, $\textsc{ExistsWeak}(\omega, \E)$ checks the existence of a trace $\pi \models \omega$ 
consistent with some $\strategyenv \in \Sigma_{\E}$. A weak strategy for $\omega$ under $\E$ exists iff $\L(\N)$ is non-empty.

The complexity of $\textsc{ExistsWeak}(\omega, \E)$ in $\omega$ is dominated by checking language non-emptiness of \mbox{$\N = \N_{\omega} \times \A'_{\E}$}. That can be done on-the-fly while building the product~\cite{BaierK08}. Being $\N_{\omega}$ exponential in $|\omega|$, we have that $\textsc{ExistsWeak}(\omega, \E)$ gives \pspace membership in $\omega$. The complexity of $\textsc{ExistsWeak}(\omega, \E)$ in $\E$ is dominated by computing and restricting the \DFA $\A_{\E}$. That can be done in doubly-exponential time in $|\E|$ and $\textsc{ExistsWeak}(\omega, \E)$ gives \twoexptime membership in $\E$. Hardness and complexity of $\existsweak(\omega, \E)$ are established in the following:

\begin{theorem}~\label{thm:cooperative-synthesis-complexity}
    \LTLf weak synthesis for $\omega$ under $\E$ is: 1. \pspace-complete in the size of $\omega$; and 2. \mbox{\twoexptime-complete} in the size of $\E$. 
\end{theorem}

We present below an algorithm to check if a strategy $\sigma_{ag}$ is winning for $\omega$ under $\E$, denoted $\checkwin(\omega, \E, \sigma_{ag})$: 1. Construct the \NFA $\N_{\lneg \omega}$ of $\lneg \omega$, the \DFA $\A_{\E}$ of $\E$, and the \DFA $\A_{\sigma_{ag}}$ of $\sigma_{ag}$; 2. Restrict $\A_\E$ to the environment winning region and obtain \DFA $\A'_{\E}$; and 3. Check language non-emptiness of the product $\N = \N_{\lneg \omega} \times \A'_{\E} \times \A_{\sigma_{ag}}$. Intuitively, $\checkwin(\omega, \E, \sigma_{ag})$ checks the existence of a trace $\pi \not \models \omega$ consistent with $\sigma_{ag}$ and some environment strategy $\sigma_{env} \in \Sigma_{\E}$, i.e., $\pi$ witnesses that $\sigma_{ag}$ is \emph{not} winning for $\omega$ under $\E$. As a result, we have that $\sigma_{ag}$ is a winning strategy for $\omega$ under $\E$ iff $\L(\N)$ is empty. 

$\checkwin(\omega, \E, \sigma_{ag})$  gives \pspace and \twoexptime membership in $\omega$ and $\E$, respectively (its complexity analysis in $\omega$ and $\E$ is analogous to that of $\existsweak(\omega, \E)$) and is polynomial in $\sigma_{ag}$, since the sizes of $\A_{\sigma_{ag}}$ and $\N$ are polynomial in that of $\sigma_{ag}$. Hardness and complexity of $\textsc{CheckWin}(\omega, \E, \sigma_{ag})$ are established in the following.



\begin{theorem}\label{thm:check-win-complexity}
    Checking if $\sigma_{ag}$ is winning for $\omega$ under $\E$ is: 1. \mbox{\pspace-complete} in the size of $\omega$; 2. \mbox{\twoexptime-complete} in the size of $\E$; and 3. Polynomial in the size of $\sigma_{ag}$.
\end{theorem}

\begin{figure}[t]
    \centering
    \includegraphics[width=.80\linewidth]{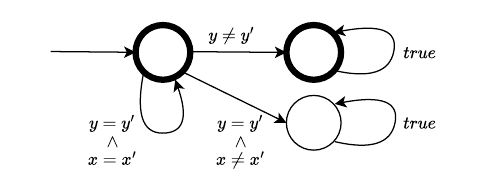}
    \caption{\DFA $\A_{\Y \neq \Y'}$. Final states are in bold. $y = y'$ (resp. $y \neq y'$) denotes equal (resp. distinct) assignments of atoms in $\Y$ and $\Y'$. Similarly for $x = x'$ (resp. $x \neq x'$).}
    \label{fig:glue-automaton}
\end{figure}

We present below an algorithm to check if a strategy $\sigma_{ag}$ is dominant for $\omega$ under $\E$, denoted $\textsc{CheckDom}(\omega, \E, \sigma_{ag})$: \begin{compactenum}
    \item Let $\Y' = \{y \tst y \in \Y\}$ and $\X' = \{x \tst x \in \X\}$;
    \item Define $\omega'$ and $\E'$ as copies of $\omega$ and $\E$ over $\Y' \cup \X'$;
    \item $\N_{\lneg \omega} = \textsc{ToNFA}(\lneg \omega)$ and $\N_{\omega'} = \textsc{ToNFA}(\omega')$
    \item $\A_{\E} = \textsc{ToDFA}(\E)$ and $\A_{\E'} = \textsc{ToDFA}(\E')$
    \item $W_{\E} = \textsc{EnvWin}(\A_\E)$ and $W_{\E'} = \textsc{EnvWin}(\A_{\E'})$
    \item $\A'_{\E} = \textsc{Restr}(\A_{\E}, W_{\E})$ and  $\A'_{\E'} = \textsc{Restr}(\A_{\E'}, W_\E')$
    \item $\A_{\sigma_{ag}} = \textsc{ToDFA}(\sigma_{ag})$
    \item $\N = \A_{\Y \neq \Y'} \times (\N_{\lneg \omega} \times \A'_{\E} \times \A_{\sigma_{ag}}) \times (\N_{\omega'}\times \A'_{\E'})$
    \item \textbf{if} $\textsc{NonEmpty}(\N)$ \textbf{return} $\false$; \textbf{else} \textbf{return} $\true$
\end{compactenum}
$\textsc{CheckDom}(\omega, \E, \sigma_{ag})$ checks the existence of two distinct traces $\pi$ and $\pi'$ such that, for the input strategy $\sigma_{ag}$ and some environment strategy $\sigma_{env}$ enforcing $\E$, we have that $\pi = \play(\sigma_{ag}, \sigma_{env}) \not \models \omega$ and, for some agent strategy $\sigma'_{ag}$ distinct from $\sigma_{ag}$, we have that $\pi' = \play(\sigma'_{ag}, \sigma_{env}) \models \omega$, i.e., $\pi$ and $\pi'$ witness the existence of a strategy $\sigma'_{ag}$ such that $\sigma_{ag} \not \geq_{\lneg \omega|\E} \sigma'_{ag}$.  It constructs the \LTLf formulas $\omega'$  and $\E'$ defined
over $\Y' \cup \X'$, where $\Y'$ and $\X'$ are ``primed” copies of $\Y$ and $\X$, and takes the product of several automata, including: (a) $\A_{\Y \neq \Y'}$, whose language is the set of pairs of traces $(\pi, \pi')$ where, if the agent makes the same choices in both $\pi$ and $\pi'$, then the environment does the same (see Figure~\ref{fig:glue-automaton}); (b) $(\N_{\lneg \omega} \times \A'_{\E} \times \A_{\sigma_{ag}})$, whose language is the set of traces $\pi=\play(\sigma_{ag}, \sigma_{env}) \not \models \omega$, where $\sigma_{ag}$ is the input strategy and $\sigma_{env}$ is some environment strategy enforcing $\E$; and (c) $(\N_{\omega'}\times \A'_{\E'})$, whose language is the set of traces $\pi' = \play(\sigma'_{ag}, \sigma_{env}) \models \omega$ for some agent strategy $\sigma'_{ag}$ and some environment strategy $\sigma_{env}$ enforcing $\E$. The language of the product automaton $\N$ is the set of pairs of traces $(\pi, \pi')$ mentioned above. It follows that $\sigma_{ag}$ is dominant for $\omega$ under $\E$ iff $\L(\N)$ is empty.
$\textsc{CheckDom}(\omega, \E, \sigma_{ag})$ gives \pspace and \twoexptime membership in the sizes of $\omega$ and $\E$, respectively, and is polynomial in the size of the strategy $\strategyag$. Hardness and complexity of $\checkdom(\omega, \E, \sigma_{ag})$  are established in the following:

\begin{theorem}~\label{thm:check-dom-complexity}
    Checking if $\sigma_{ag}$ is dominant for $\omega$ under $\E$ is: 1. {\pspace-complete} in the size of $\omega$; 2. {\twoexptime-complete} in the size of $\E$; and 3. Polynomial in the size of $\sigma_{ag}$.  
\end{theorem}


Checking if a strategy $\sigma_{ag}$ is best-effort for $\omega$ under $\E$ is computationally harder than checking winning and dominant strategies. It requires a sophisticated game-theoretic technique that extends the best-effort synthesis algorithm in~\cite{AminofDR21}. We present below our technique, denoted $\checkbe(\omega, \E, \sigma_{ag})$: 
\begin{compactenum}
    \item $\A_{\omega} = \textsc{ToDFA}(\omega)$ and $\A_{\E} = \textsc{ToDFA}(\E)$;
    \item $W_{\E} = \textsc{EnvWin}(\A_{\E})$; and $\A'_{\E} = \textsc{Restr}(\A_\E, W_\E)$;
    \item $\G = \A_{\omega} \times \A'_{\E}$;
    \item $W = \textsc{WinRegion}(\G)$ and $W' = \textsc{WeakRegion}(\G)$;
    \item $\A_{\sigma_{ag}} = \textsc{ToDFA}(\sigma_{ag})$
    \item $\G' = \G \times \A_{\sigma_{ag}}$.Say: $S_{\G} \times S_{\sigma_{ag}}$ is the state set of $\G'$; $s_0$ is the initial state of $\G'$; $F$ is the set of final states of $\G'$;
    \item $W_{\G'} = \{(s_\G, s_{\sigma_{ag}}) \in S_\G \times S_{\sigma_{ag}}) \mid s_{\G} \in W\}$;
    \item $W'_{\G'} = \{(s_\G, s_{\sigma_{ag}}) \in S_\G \times S_{\sigma_{ag}}) \mid s_{\G} \in W'\}$;
    \item \textbf{for} every state $s \in S_{\G} \times S_{\sigma_{ag}}$ reachable from $s_0$: \begin{compactitem}
        \item \textbf{if} $s \in W_{\G'}$ and there exists a path $\rho = s \cdots s'$ such that $s' \not \in F$ and $\rho$ can't be extended to reach $F$ \mbox{\textbf{return} \textit{false}};
        \item \textbf{else if} $s \in W'_{\G'}\setminus W_{\G'}$ and no path from $s$ to $F$ exists \textbf{return} \textit{false}; 
    \end{compactitem}
    \item \textbf{return} \textit{true} 
\end{compactenum}

$\checkbe(\omega, \E, \sigma_{ag})$ uses the local characterization of best-effort strategies in Preliminaries to check whether $\sigma_{ag}$ is best-effort for $\omega$ under $\E$ or not. It solves \DFA games obtained from $\omega$ and $\E$~(Lines 1-4) with the following property. Histories $h$ whose induced runs lead to a state $s$ in the winning region $W$ (resp. weakly winning region $W' \setminus W$) achieves  $val_{\omega|\E}(h) = +1$ (resp. $val_{\omega|\E}(h) = 0$)~\cite{AminofDR21}. To evaluate the value that $\sigma_{ag}$ achieves in each history consistent with it, $\checkbe(\omega, \E, \sigma_{ag})$ builds the product $\G' = \G \times \A_{\sigma_{ag}}$ (Lines 5 and 6) and lifts the regions $W$ and $W'$ to $\G'$ to obtain the regions $W_{\G'}$ and $W'_{\G'}$~(Lines 7 and 8). For every state in $W_{\G'}$ (resp. \mbox{$W'_{\G'} \setminus W_{\G'}$}), $\checkbe(\omega, \E)$ checks that every execution (resp. the existence of an execution) of $\sigma_{ag}$ leads to a final state, so that $val_{\omega|\E}(\sigma_{ag}, h) = +1$ (resp. $val_{\omega|\E}(\sigma_{ag}, h) = 0$) (Line 9). If not, there exists a history $h$ such that \mbox{$val_{\omega|\E}(\sigma_{ag}, h) < val_{\omega|\E}(h)$}. That violates the local characterization of best-effort strategies and $\sigma_{ag}$ is \emph{not} best-effort for $\omega$ under $\E$. \mbox{Else, $\sigma_{ag}$ is best-effort for $\omega$ under $\E$. (Line 10)} 

The complexity of $\checkbe(\omega, \E, \sigma_{ag})$ in $\omega$ and $\E$ is dominated by constructing and solving \DFA games that are doubly-exponential in $|\omega|$ and $|\E|$ \cite{DegVa15}, i.e., $\checkbe(\omega, \E, \sigma_{ag})$ gives \twoexptime membership in $\omega$ and $\E$. The complexity of $\checkbe(\omega, \E, \sigma_{ag})$ is polynomial in $\sigma_{ag}$ as the sizes of $\A_{\sigma_{ag}}$ and $\G'$ are polynomial in that of $\sigma_{ag}$. Hardness and complexity of $\checkbe(\omega, \E, \sigma_{ag})$ are established in the following.

\begin{theorem}\label{thm:best-effort-checking-complexity}
    Checking if $\sigma_{ag}$ is best-effort for $\omega$ under $\E$ is: 1. \mbox{\twoexptime-complete} in the sizes of $\omega$ and $\E$; and 2. Polynomial in the size of $\sigma_{ag}$.
\end{theorem}

\begin{figure}[t]
    \centering \includegraphics[width=0.99\linewidth]{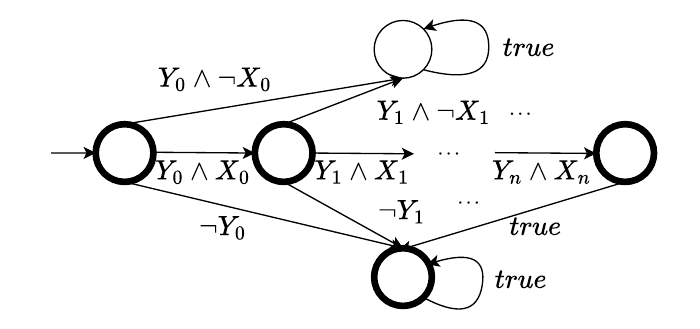}
    \caption{\DFA $\A_{\E_h}$ of \LTLf environment specification $\E_h$ that captures environment strategies $\sigma_{env}$ such that history $h = (Y_0 \cup X_0) \cdots (Y_n \cup X_n)$ is consistent with $\sigma_{env}$.} 
    \label{fig:history-dfa}
\end{figure}

When the environment specification is given as $\E \land \E_h$, where $\E_h$ captures environment strategies consistent with a history $h = (Y_0 \cup X_0) \cdots (Y_n \cup X_n)$, \DFA $\A_{\E \land \E_h}$ can be built as $\A_{\E} \times \A_{\E_h}$. By observing that \DFA $\A_{\E_h}$ can be built in polynomial time in the length of $h$ as in Figure~\ref{fig:history-dfa} (note that transitions of the form $Y_i \land \lneg X_i$ lead to a non-final sink state, as no environment strategy $\sigma_{env}$ such that $h$ is consistent with $\sigma_{env}$ plays  $\lneg X_i$ when the agent plays $Y_0 \cdots Y_i$), we have that the checking techniques above requires polynomial time in $h$. That also explains the polynomial complexity in $h$ of the responsibility checking algorithms in Table~\ref{table:responsibility-checking}.

Table~\ref{table:responsibility-checking} does not mention passive and inexcusable passive responsibility attribution against environment strategies. This is because such a computational grounding requires restricting and representing sets of environment strategies. If environment strategies can be represented as finite-state transducers, we can adopt the tree-automata techniques in~\cite{AminofDLMR20,AminofDLMR21} to computationally characterize these notions. These techniques establish decidability of responsibility attribution against environment strategies, but not tight complexity bounds. As a result, we leave the computational grounding of passive responsibility attribution against environment strategies to future work.

\section{Conclusions}\label{sec:conclusion}

We gave a computational grounding to notions of responsibility attribution and anticipation in an \LTLf strategy-based setting. In particular it comes as a welcome surprise that such notions are related to notions independently developed for best-effort synthesis.
While our investigation is theoretical, it is possible to implement responsibility checking, e.g., using game-theoretic techniques~\cite{fijalkow2023gamesgraphs}. 


Besides checking, we can also synthesize strategies that do not anticipate passive or inexcusable passive responsibility.
Similarly we can synthesize strategies that do (or do not) anticipate active responsibility. Based on Theorems~\ref{thm:pass-and-dom}  
and~\ref{thm:pass-and-dom-2}, such synthesis tasks are reducible (in polynomial time) to dominant, best-effort, and reactive synthesis~\cite{DegVa15,AminofDMR19,AminofDR21,AminofDR2023}. As a result, synthesizing strategies that are responsibility aware is \twoexptime-complete.

Our responsibility analysis takes a first-person perspective from the viewpoint of the agent (vs. a third-person perspective from the viewpoint of the designer). If we want to extend it to a multi-agent setting, each agent will have a different environment formed by the environment itself and the agent's peers acting in it. We plan to study this in the future.

\bibliography{aaai25}

\begin{thebibliography}{33}
\providecommand{\natexlab}[1]{#1}

\bibitem[{Abarca and Broersen(2022)}]{DBLP:conf/atal/AbarcaB22}
Abarca, A. I.~R.; and Broersen, J.~M. 2022.
\newblock A Stit Logic of Responsibility.
\newblock In \emph{{AAMAS}}, 1717--1719. International Foundation for Autonomous Agents and Multiagent Systems {(IFAAMAS)}.

\bibitem[{Aminof et~al.(2020)Aminof, {De Giacomo}, Lomuscio, Murano, and Rubin}]{AminofDLMR20}
Aminof, B.; {De Giacomo}, G.; Lomuscio, A.; Murano, A.; and Rubin, S. 2020.
\newblock Synthesizing Strategies under Expected and Exceptional Environment Behaviors.
\newblock In \emph{{IJCAI}}, 1674--1680.

\bibitem[{Aminof et~al.(2021)Aminof, {De Giacomo}, Lomuscio, Murano, and Rubin}]{AminofDLMR21}
Aminof, B.; {De Giacomo}, G.; Lomuscio, A.; Murano, A.; and Rubin, S. 2021.
\newblock Synthesizing Best-effort Strategies under Multiple Environment Specifications.
\newblock In \emph{{KR}}, 42--51.

\bibitem[{Aminof et~al.(2019)Aminof, {De Giacomo}, Murano, and Rubin}]{AminofDMR19}
Aminof, B.; {De Giacomo}, G.; Murano, A.; and Rubin, S. 2019.
\newblock Planning under {LTL} Environment Specifications.
\newblock In \emph{{ICAPS}}, 31--39.

\bibitem[{Aminof, {De Giacomo}, and Rubin(2021)}]{AminofDR21}
Aminof, B.; {De Giacomo}, G.; and Rubin, S. 2021.
\newblock Best-Effort Synthesis: Doing Your Best Is Not Harder Than Giving Up.
\newblock In \emph{{IJCAI}}, 1766--1772.

\bibitem[{Aminof, {De Giacomo}, and Rubin(2023)}]{AminofDR2023}
Aminof, B.; {De Giacomo}, G.; and Rubin, S. 2023.
\newblock Reactive Synthesis of Dominant Strategies.
\newblock In \emph{{AAAI}}, 6228--6235.

\bibitem[{Apt and Gr{\"{a}}del(2011)}]{AptG2011}
Apt, K.~R.; and Gr{\"{a}}del, E., eds. 2011.
\newblock \emph{Lectures in Game Theory for Computer Scientists}.
\newblock Cambridge University Press.

\bibitem[{Baier, Funke, and Majumdar(2021)}]{Baier0M21}
Baier, C.; Funke, F.; and Majumdar, R. 2021.
\newblock A Game-Theoretic Account of Responsibility Allocation.
\newblock In \emph{{IJCAI}}, 1773--1779.

\bibitem[{Baier and Katoen(2008)}]{BaierK08}
Baier, C.; and Katoen, J. 2008.
\newblock \emph{Principles of Model Checking}.
\newblock {MIT} Press.

\bibitem[{Baltag, Canavotto, and Smets(2021)}]{Baltag2021-BALCAA-8}
Baltag, A.; Canavotto, I.; and Smets, S. 2021.
\newblock Causal Agency and Responsibility: A Refinement of {STIT} Logic.
\newblock In Giordani, A.; and Malinowski, J., eds., \emph{Logic in High Definition, Trends in Logical Semantics}, 149--176.

\bibitem[{Bansal et~al.(2023)Bansal, Li, Tabajara, Vardi, and Wells}]{BansalLTVW23}
Bansal, S.; Li, Y.; Tabajara, L.~M.; Vardi, M.~Y.; and Wells, A.~M. 2023.
\newblock Model Checking Strategies from Synthesis over Finite Traces.
\newblock In \emph{{ATVA} {(1)}}, volume 14215 of \emph{Lecture Notes in Computer Science}, 227--247. Springer.

\bibitem[{Belnap, Perloff, and Xu(2001)}]{belnap01facing}
Belnap, N.; Perloff, M.; and Xu, M. 2001.
\newblock \emph{Facing the Tuture: Agents and Choices in our Indeterminist World}.
\newblock New York: Oxford University Press.

\bibitem[{Braham and van Hees(2012)}]{Braham2012}
Braham, M.; and van Hees, M. 2012.
\newblock An Anatomy of Moral Responsibility.
\newblock \emph{Mind}, 121(483): 601--634.

\bibitem[{Bulling and Dastani(2013)}]{DBLP:conf/clima/BullingD13}
Bulling, N.; and Dastani, M. 2013.
\newblock Coalitional Responsibility in Strategic Settings.
\newblock In \emph{{CLIMA}}, Lecture Notes in Computer Science, 172--189.

\bibitem[{Chockler and Halpern(2004)}]{DBLP:journals/jair/ChocklerH04}
Chockler, H.; and Halpern, J. 2004.
\newblock Responsibility and Blame: {A} Structural-Model Approach.
\newblock \emph{Journal of Artificial Intelligence Research}, 22: 93--115.

\bibitem[{Cimatti et~al.(2003)Cimatti, Pistore, Roveri, and Traverso}]{CimattiPRT03}
Cimatti, A.; Pistore, M.; Roveri, M.; and Traverso, P. 2003.
\newblock Weak, Strong, and Strong Cyclic Planning via Symbolic Model Checking.
\newblock \emph{AIJ}, 1--2(147): 35--84.

\bibitem[{{De Giacomo} et~al.(2021){De Giacomo}, {Di Stasio}, Perelli, and Zhu}]{DeGiacomoDPZ21}
{De Giacomo}, G.; {Di Stasio}, A.; Perelli, G.; and Zhu, S. 2021.
\newblock Synthesis with Mandatory Stop Actions.
\newblock In \emph{{KR}}, 237--246.

\bibitem[{{De Giacomo} and Vardi(2013)}]{DegVa13}
{De Giacomo}, G.; and Vardi, M.~Y. 2013.
\newblock Linear Temporal Logic and Linear Dynamic Logic on Finite Traces.
\newblock In \emph{{IJCAI}}, 854--860.

\bibitem[{{De Giacomo} and Vardi(2015)}]{DegVa15}
{De Giacomo}, G.; and Vardi, M.~Y. 2015.
\newblock Synthesis for {LTL} and {LDL} on Finite Traces.
\newblock In \emph{{IJCAI}}, 854--860.

\bibitem[{Fijalkow et~al.(2023)Fijalkow, Bertrand, Bouyer-Decitre, Brenguier, Carayol, Fearnley, Gimbert, Horn, Ibsen-Jensen, Markey, Monmege, Novotný, Randour, Sankur, Schmitz, Serre, and Skomra}]{fijalkow2023gamesgraphs}
Fijalkow, N.; Bertrand, N.; Bouyer-Decitre, P.; Brenguier, R.; Carayol, A.; Fearnley, J.; Gimbert, H.; Horn, F.; Ibsen-Jensen, R.; Markey, N.; Monmege, B.; Novotný, P.; Randour, M.; Sankur, O.; Schmitz, S.; Serre, O.; and Skomra, M. 2023.
\newblock Games on Graphs.
\newblock arXiv:2305.10546.

\bibitem[{Gale and Stewart(1953)}]{gale1953infinite}
Gale, D.; and Stewart, F.~M. 1953.
\newblock Infinite Games with Perfect Information.
\newblock \emph{Contributions to the Theory of Games}, 2(245-266): 2--16.

\bibitem[{Jansen(2014)}]{Jansen14}
Jansen, N. 2014.
\newblock The Idea of Legal Responsibility.
\newblock \emph{Oxford Journal of Legal Studies}, 34(2): 221--252.

\bibitem[{Lorini, Longin, and Mayor(2014)}]{DBLP:journals/logcom/LoriniLM14}
Lorini, E.; Longin, D.; and Mayor, E. 2014.
\newblock A Logical Analysis of Responsibility Attribution: Emotions, Individuals and Collectives.
\newblock \emph{Journal of Logic and Computation}, 24(6): 1313--1339.

\bibitem[{Lorini and M{\"{u}}hlenbernd(2018)}]{DBLP:journals/fuin/LoriniM18}
Lorini, E.; and M{\"{u}}hlenbernd, R. 2018.
\newblock The Long-Term Benefits of Following Fairness Norms under Dynamics of Learning and Evolution.
\newblock \emph{Fundamenta Informaticae}, 158(1-3): 121--148.

\bibitem[{Lorini and Schwarzentruber(2011)}]{DBLP:journals/ai/LoriniS11}
Lorini, E.; and Schwarzentruber, F. 2011.
\newblock A Logic for Reasoning about Counterfactual Emotions.
\newblock \emph{Artificial Intelligence}, 175(3-4): 814--847.

\bibitem[{Naumov and Tao(2021)}]{DBLP:conf/ijcai/Naumov021}
Naumov, P.; and Tao, J. 2021.
\newblock Two Forms of Responsibility in Strategic Games.
\newblock In \emph{{IJCAI}}, 1989--1995.

\bibitem[{Naumov and Tao(2023)}]{DBLP:journals/apal/NaumovT23}
Naumov, P.; and Tao, J. 2023.
\newblock Counterfactual and Seeing-To-It Responsibilities in Strategic Games.
\newblock \emph{Annals of Pure and Applied Logic}, 174(10): 103353.

\bibitem[{Parker, Grandi, and Lorini(2023)}]{DBLP:conf/ecai/ParkerGL23}
Parker, T.; Grandi, U.; and Lorini, E. 2023.
\newblock Anticipating Responsibility in Multiagent Planning.
\newblock In \emph{{ECAI}}, 1859--1866.

\bibitem[{Pnueli(1977)}]{Pnu77}
Pnueli, A. 1977.
\newblock The Temporal Logic of Programs.
\newblock In \emph{{FOCS}}, 46--57.

\bibitem[{Shi(2024)}]{DBLP:conf/aaai/Shi24}
Shi, Q. 2024.
\newblock Responsibility in Extensive Form Games.
\newblock In \emph{{AAAI}}, 19920--19928.

\bibitem[{Talbert(2023)}]{sep-moral-responsibility}
Talbert, M. 2023.
\newblock {Moral Responsibility}.
\newblock In Zalta, E.~N.; and Nodelman, U., eds., \emph{The {Stanford} Encyclopedia of Philosophy}. Metaphysics Research Lab, Stanford University, {F}all 2023 edition.

\bibitem[{Vincent(2011)}]{Vincent2011}
Vincent, N.~A. 2011.
\newblock \emph{A Structured Taxonomy of Responsibility Concepts}, 15--35.
\newblock Dordrecht: Springer Netherlands.

\bibitem[{Yazdanpanah et~al.(2019)Yazdanpanah, Dastani, Jamroga, Alechinsa, and Logan}]{DBLP:conf/atal/YazdanpanahDJAL19}
Yazdanpanah, V.; Dastani, M.; Jamroga, W.; Alechinsa, N.; and Logan, B. 2019.
\newblock Strategic Responsibility Under Imperfect Information.
\newblock In \emph{AAMAS 2019}, 592--600.

\end{thebibliography}

\newpage
\section{Supplementary Material}\label{sec:supplementary}
We first give hardness results for the problems of checking winning, dominant, and best-effort strategies and the existence of weak strategies. That completes the proofs of Theorems~\ref{thm:cooperative-synthesis-complexity},~\ref{thm:check-win-complexity},~\ref{thm:check-dom-complexity}, and~\ref{thm:best-effort-checking-complexity}. Subsequently, we give complete proofs for the results showing the connection between active, passive, and inexcusable passive responsibility with winning, dominant, and best-effort strategies, respectively, i.e., Theorems~\ref{thm:pass-and-dom} and~\ref{thm:pass-and-dom-2}. We conclude by giving complete proofs for the results of our computational grounding, i.e., Theorem~\ref{thm:responsibility-checking-thm}.

\subsection{Hardness Proofs}


\setcounter{theorem}{3}
\begin{theorem}~\label{thm:cooperative-synthesis-complexity}
    \LTLf weak synthesis for $\omega$ under $\E$ is: \begin{compactitem}
        \item \pspace-complete in the size of $\omega$; 
        \item \twoexptime-complete in the size of $\E$.
    \end{compactitem}
\end{theorem}

\begin{proof}
    \pspace-hardness follows immediately by reducing from \LTLf satisfability, which is \pspace-hard~\cite{DegVa13}.
    \twoexptime-hardness follows by reducing from checking environment enforceability of an \LTLf formula $\E$, i.e., deciding the existence of an environment strategy enforcing $\E$, which is \twoexptime-hard~\cite{AminofDMR19}.
    We show that $\E$ is environment enforceable iff there exists a weak strategy for $\lneg x \land y$ under $\E \vee x$, where $y$ (resp. $x$) is a new atom under agent's (resp. environment's) control. Assume that $\E$ is environment enforceable. Let $\sigma_{env}$ be an environment strategy enforcing $\E$. Observe that $\sigma_{env}$ enforces $\E \vee x$ and never plays $x$. Let $\sigma_{ag}$ be the agent strategy such that $\sigma_{ag}(\lambda) = \{y\}$ and $\sigma_{ag}(X) = \stop$ for every other $X \in (2^{\X})^*$. We have that $\play(\sigma_{ag}, \sigma_{env}) \models \lneg x \land y$ and hence $\sigma_{ag}$ is weak for $\lneg x \land y$ under $\E \lor x$. 
    Conversely, assume that $\E$ is not environment enforceable. Then, $\E \vee x$ is environment enforceable only by environment strategies that play $x$ in the first time step. Hence, $\lneg x \land y$ cannot be satisfied and no weak strategy for $\lneg x \land y$ under $\E \lor x$  exists.
\end{proof}

We now turn to the hardness proof of Theorem~\ref{thm:check-win-complexity}, i.e., hardness of checking winning strategies. As a preliminary step, we investigate the problem of checking weak strategies, i.e., deciding if an input agent strategy given in the form of a terminating transducer is weak or not. We will then prove the hardness of checking winning strategies by reducing from checking weak strategies. 

We present below an algorithm to check if a strategy is weak, denoted $\checkweak(\omega, \E)$: \begin{compactenum}
    \item $\N_{\omega} = \textsc{ToNFA}(\omega)$
    \item $\A_{\E} = \textsc{ToDFA}(\E)$ 
    \item $W_{\E} = \textsc{EnvWin}(\A_\E)$
    \item $\A'_{\E} = \textsc{Restr}(\A_{\E}, W_\E)$
    \item $\A_{\sigma_{ag}} = \textsc{ToDFA}(\sigma_{ag})$ 
    \item $\N = \N_{\omega} \times \A'_{\E} \times \A'_{\sigma_{ag}}$
    \item \textbf{if} $\textsc{NonEmpty}(\N)$ \textbf{return} \textit{true}; \textbf{else} \textbf{return} \textit{false}
\end{compactenum}
$\textsc{CheckWeak}(\omega, \E, \sigma_{ag})$ checks if there exists a trace $\pi$ that satisfies $\omega$ and is consistent with $\sigma_{ag}$ and some environment strategy enforcing $\E$. If such a trace exists (resp. does not exist), $\sigma_{ag}$ is (resp. is not) a weak strategy.

The complexity analysis of $\textsc{CheckWeak}(\omega, \E, \sigma_{ag})$ in $\omega$ and $\E$ is analogous to that of $\textsc{ExistsWeak}(\omega, \E)$, giving \pspace and \twoexptime membership in $\omega$ and $\E$, respectively. The product \mbox{$\N = \N_{\omega} \times \A'_{\E} \times \A_{\sigma_{ag}}$} is polynomial in the size of $\sigma_{ag}$, i.e., checking non-emptiness of $\L(\N)$ is polynomial in the size of $\sigma_{ag}$. Hardness and complexity of checking weak strategies are established in the following:

\setcounter{theorem}{7}
\begin{theorem}\label{thm:check-coop-complexity}
    \mbox{Checking if $\sigma_{ag}$ is weak for $\omega$ under $\E$ is:} \begin{compactitem}
    \item \pspace-complete in the size of $\omega$;
    \item \twoexptime-complete in the size of $\E$;
    \item Polynomial in the size of $\sigma_{ag}$.
    \end{compactitem}
\end{theorem}

\begin{proof}[Proof.]
   \pspace-hardness follows by reducing from checking if $\sigma_{ag}$ is winning for $\omega$ without environment, which is \pspace-hard if $\sigma_{ag}$ is represented as a terminating transducer~\cite{BansalLTVW23}, by observing
    that $\sigma_{ag}$ is winning for $\omega$ iff $\sigma_{ag}$ is not weak for $\lneg \omega$. \twoexptime-hardness follows by reducing from checking environment enforceability of $\E$~\cite{AminofDMR19} as in Theorem~\ref{thm:cooperative-synthesis-complexity}. 
\end{proof}

We now prove hardness of checking winning strategies. 

\setcounter{theorem}{4}
\begin{theorem}\label{thm:check-win-complexity}
    \mbox{Checking if $\sigma_{ag}$ is winning for $\omega$ under $\E$ is:}\begin{compactitem}
        \item \pspace-complete in the size of $\omega$;
        \item \twoexptime-complete in the size of $\E$;
        \item Polynomial in the size of $\sigma_{ag}$.
    \end{compactitem}
\end{theorem}

\begin{proof}
    The claim follows immediately by reducing from checking weak strategies and observing that $\sigma_{ag}$ is winning for $\omega$ under $\E$ iff $\sigma_{ag}$ is not weak for $\lneg \omega$ under $\E$. 
\end{proof}

We now turn to the hardness proofs of Theorems~\ref{thm:check-dom-complexity} and~\ref{thm:best-effort-checking-complexity}, i.e., hardness of checking dominant and best-effort strategies, respectively. Both proofs use the local characterization of dominant and best-effort strategies in Preliminaries. 

\begin{theorem}~\label{thm:check-dom-complexity}
    Checking if $\sigma_{ag}$ is dominant for $\omega$ under $\E$: \begin{compactitem}
        \item \pspace-complete in the size of $\omega$;
        \item \twoexptime-complete in the size of $\E$;
        \item Polynomial in the size of $\sigma_{ag}$.
    \end{compactitem}
\end{theorem} \begin{proof}
    We prove \pspace- and \twoexptime-hardness by reducing from checking winning strategies, see Theorem~\ref{thm:check-win-complexity}. We show that $\sigma_{ag}$ is winning for $\varphi$ (under $\E$) iff $\sigma_{ag}$ is dominant for $\varphi' = \varphi \lor ((y \lor y') \land x)$ (under $\E$), where $y$ and $y'$ (resp. $x$) are new atoms under agent's (resp. environment's) control. Assume that $\sigma_{ag}$ is winning for $\varphi$. We have that $\sigma_{ag}$ is also winning (i.e., dominant) for $\varphi'$. Conversely, assume that $\sigma_{ag}$ is not winning for $\varphi$. There are two possibilities: \myi there exists a strategy that is winning for $\varphi$; or \myii no winning strategy for $\varphi$ exists. Suppose \myi holds. We have that $\sigma_{ag}$ is not dominant for $\varphi'$ since it is not winning for $\varphi$. Suppose that \myii holds. We have that $val_{\varphi'|\E}(\lambda) = 0$, since the agent can win (resp. lose) $\varphi'$ by playing $y$ in the first time step together with the environment strategy that plays (resp. does not play) $x$ in the first time step. However, observe that there exist at least two agent moves $Y \neq \sigma_{ag}(\lambda)$, i.e., $\{y\}$ and $\{y'\}$, such that $val_{\varphi'|\E}(\lambda) = 0$ and $val_{\varphi'|\E}(\lambda \cdot Y) = 0$. That violates the local characterization of dominant strategies in Preliminaries and $\sigma_{ag}$ is not dominant for $\varphi'$.
\end{proof}

\begin{theorem}\label{thm:best-effort-checking-complexity}
    Checking if $\sigma_{ag}$ is best-effort for $\omega$ under $\E$ is\begin{compactitem}
        \item \twoexptime-complete in the sizes of $\omega$ and $\E$;
        \item Polynomial in the size of $\sigma_{ag}$
    \end{compactitem}
\end{theorem}
\begin{proof}
    We prove hardness by reducing from \LTLf reactive synthesis (without environment specifications), which is \twoexptime-hard~\cite{DegVa15}. We show that there exists a winning strategy for $\varphi$ iff $\sigma_{ag}$ such that $\sigma_{ag}(\lambda) = \{y\}$  and \mbox{$\sigma_{ag}(X) = \stop$} for every other $X \in (2^{\X})^*$ is \emph{not} best-effort for \mbox{$\varphi' = \varphi \oplus (y \land x)$}, where $y$ (resp. $x$) is a new atom under agent's (resp. environment's) control and $\oplus$ is exclusive-or. Assume that there exists a winning strategy for $\varphi$. Then $\sigma_{ag}$ is not winning (i.e., not best-effort) for $\varphi'$ as it is not winning for $\varphi$. Conversely, assume that no winning strategy for $\varphi$ exists. Observe that $val_{\varphi'}(\lambda) = 0$ and $val_{\varphi'}(\sigma_{ag}, \lambda) = 0$, as $\sigma_{ag}$ satisfies (resp. does not satisfy) $\varphi'$ together with the environment strategy that plays (resp. does not play) $x$ in the first time step. With this  result, and observing that $\sigma_{ag}$ stops after the first time step, 
    we have that $\sigma_{ag}$ satisfies the local characterization of best-effort strategies in Preliminaries and is best-effort for $\omega$. Hardness wrt the environment can be proved by reducing from checking best-effort strategies without environment specifications.
\end{proof}

\subsection{Proof of Theorems~\ref{thm:pass-and-dom},~\ref{thm:pass-and-dom-2}, and~\ref{thm:responsibility-checking-thm}}


\setcounter{theorem}{0}
\begin{theorem}\label{thm:pass-and-dom}~
\begin{compactitem}
        \item The agent anticipates passive (resp. inexcusable passive) responsibility for $\omega$ under $\strategyag$ and $\E$ iff $\strategyag$ is not dominant (resp. is not best-effort) for $\lnot \omega$ under $\E$.
        \item The agent anticipates active responsibility for $\omega$ under $\strategyag$ and $\E$ iff $\strategyag$ is winning for $\omega$ under $\E$ and there exists some $\strategyag'$ that is weak for $\lneg \omega$ under $\E$;
    \end{compactitem}
\end{theorem}

\begin{proof}
    The agent anticipates passive responsibility for $\omega$ under $\sigma_{ag}$ and $\E$ iff there is an environment strategy $\sigma_{env} \in \Sigma_{\E}$ such that: \myi $\play(\sigma_{ag},\sigma_{env}) \models \omega$; and \myii there is an agent strategy $\sigma'_{ag}$ such that $\play(\sigma'_{ag}, \sigma_{env}) \models \lneg \omega$, i.e., $\sigma_{ag} \not \geq_{\lneg \omega|\E} \sigma'_{ag}$ and $\sigma_{ag}$ is not dominant for $\lneg \omega$ under $\E$. 

    The agent anticipates inexcusable passive responsibility for $\omega$ under $\sigma_{ag}$ and $\E$ iff there is an environment strategy $\sigma_{env} \in \Sigma_{\E}$ such that: \myi $\play(\sigma_{ag}\sigma_{env}) \models \omega$; and \myii there is an agent strategy $\sigma'_{ag}$ such that $\sigma'_{ag} \geq_{\lneg \omega|\E} \sigma_{ag}$ and $\play(\sigma'_{ag}, \sigma_{env}) \models \lneg \omega$, i.e., $\sigma'_{ag} >_{\lneg \omega|\E} \sigma_{ag}$ and $\sigma_{ag}$ is not best-effort for $\lneg \omega$ under $\E$.


    The agent anticipates active responsibility for $\omega$ under $\sigma_{ag}$ and $\E$ iff: \myi \mbox{$\play(\sigma_{ag},\sigma_{env}) \models \omega$} for every environment strategy $\sigma_{env} \in \Sigma_{\E}$, i.e., $\sigma_{ag}$ is winning for $\omega$ under $\E$; and \myii there is a pair of strategies $\sigma'_{ag}$ and $\sigma'_{env} \in \Sigma_{\E}$ such that \mbox{$\play(\sigma'_{ag},\sigma'_{env}) \models \lneg \omega$}, i.e., there exists an agent strategy $\sigma'_{ag}$ that is weak for $\lneg \omega$ under $\E$.
\end{proof}

For the proof of Theorem~\ref{thm:pass-and-dom-2}, recall that for a history \mbox{$h = (Y_0 \cup X_0) \cdots (Y_n \cup X_n)$} we can construct an \LTLf environment specification $\E_h$ that captures all environment strategies such that $h$ is consistent with them as follows:
$$
\begin{array}{l}
\E_h = (Y_0 \limp X_0) \land ((Y_0 \land \Wnext^1 Y_1) \limp \Wnext^1 X_1)~\land {}\\
\cdots~\land~((Y_0 \land \Wnext^1 Y_1 \land \cdots \land \Wnext^n Y_n) \limp \Wnext^n X_n)
\end{array}
$$

\begin{theorem}\label{thm:pass-and-dom-2}
    The agent is attributed passive (resp. inexcusable passive) responsibility for $\omega$ under $\strategyag$, $\E$, and $h$ iff $\strategyag$ is not dominant (resp. best-effort) for $\lneg \omega$ under $\E \land \E_h$.
\end{theorem}

\begin{proof}
    Note that $h$ is consistent with an environment strategy $\sigma_{env}$ iff $\sigma_{env} \in \Sigma_{\E_h}$, i.e., $\sigma_{env}$ enforces $\E_h$. It follows that $\E \land \E_h$ is enforced by all environment strategies $\sigma_{env}$ enforcing $\E$ and such that $h$ is consistent with $\sigma_{env}$. With this result, the claim can be proved as shown in Theorem~\ref{thm:pass-and-dom}.
\end{proof}

\begin{table}[t]
    \centering
    \resizebox{.99\linewidth}{!}
    {\begin{tabular}{|l||l|l|}
    \hline
        &\textbf{Complexity}$(\omega, \E, \sigma_{ag})$&\textbf{Algorithm$(\omega, \E, \sigma_{ag})$}\\
    \hline \hline
    \textsc{PRAnt} & \makecell[l]{\pspace-C $(\omega)$ \\
    \twoexptime-C $(\E)$\\ 
    poly $(\sigma_{ag})$} & \makecell[l]{$\lneg \checkdom(\lneg \omega, \E, \strategyag)$} \\ \hline
    \makecell[l]{\textsc{IPRAnt}} & \makecell[l]{\twoexptime-C [$\omega$][$\E$] \\ poly $(\sigma_{ag})$} & \makecell[l]{$\lneg \checkbe(\lneg \omega, \E, \sigma_{ag})$} \\ \hline
    \textsc{PRattr}$(h)$ & \makecell[l]{\pspace-C 
 $(\omega)$ \\
    \twoexptime-C $(\E)$\\ 
    poly $[\sigma_{ag}][h]$} & \makecell[l]{$\lneg \checkdom(\lneg \omega, \E \land \E_h, \sigma_{ag})$} \\ \hline
    \textsc{IPRattr}$(h)$ & \makecell[l]{
    \twoexptime-C $[\omega][\E]$\\ 
    poly $[\sigma_{ag}][h]$} & \makecell[l]{$\lneg \checkbe(\lneg \omega, \E \land \E_h, \sigma_{ag})$}\\ \hline
    \textsc{ARA} & \makecell[l]{\pspace-C 
 $(\omega)$\\
    \twoexptime-C $(\E)$\\ 
    poly $(\sigma_{ag})$} & \makecell[l]{$\checkwin(\omega, \E, \strategyag) \land{}$\\$\textsc{ExistsWeak}(\lneg \omega, \E)$} \\ \hline
    \end{tabular}}
    \caption{Computational grounding (complexity and algorithm) of passive and active responsibility anticipation and passive responsibility attribution on histories.}
    \label{table:responsibility-checking}
\end{table}

\setcounter{figure}{+1}
\begin{figure}[t]
    \centering \includegraphics[width=0.99\linewidth]{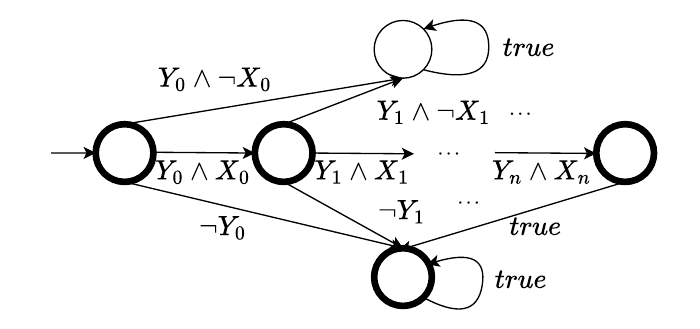}
    \caption{\DFA $\A_{\E_h}$ of \LTLf environment specification $\E_h$ that captures environment strategies $\sigma_{env}$ such that history $h = (Y_0 \cup X_0) \cdots (Y_n \cup X_n)$ is consistent with $\sigma_{env}$.} 
    \label{fig:history-dfa}
\end{figure}

\begin{theorem}\label{thm:responsibility-checking-thm}
    The worst-case computational complexity of the various responsibility notions is established in Table~\ref{table:responsibility-checking}.
\end{theorem}

\begin{proof}
    By Theorem~\ref{thm:pass-and-dom}, \textsc{PRAnt} and \textsc{IPRAnt} are interreducible with checking dominant and best-effort strategies, respectively. Then, we have by Theorem~\ref{thm:check-dom-complexity} that \textsc{PRAnt} is \pspace- and \twoexptime-complete in $|\omega|$ and $|\E|$, respectively, and by Theorem~\ref{thm:best-effort-checking-complexity} that \textsc{IPRAnt} is \twoexptime-complete in both $|\omega|$ and $|\E|$. Polynomial complexity of \textsc{PRAnt} and \textsc{IPRAnt} in $\sigma_{ag}$ follows as $\lneg \textsc{CheckDom}(\lneg \omega, \E, \sigma_{ag})$ and $\lneg \textsc{CheckBe}(\lneg \omega, \E, \sigma_{ag})$ run in polynomial time in $|\sigma_{ag}|$. 
    
    By Theorem~\ref{thm:pass-and-dom-2} we have analogous results for $\textsc{PRAttr}(h)$ and $\textsc{IPRAttr}(h)$. Polynomial complexity of $\textsc{PRAttr}(h)$ and $\textsc{IPRAttr}(h)$ in $h$ follows as \DFA $\A_{\E_h}$ can be constructed in polynomial time in $|h|$ as in Figure~\ref{fig:history-dfa}. That enables \mbox{$\lneg \textsc{CheckDom}(\lneg \omega, \E \land \E_h, \sigma_{ag})$} and \mbox{$\lneg \textsc{CheckBe}(\lneg \omega, \E \land \E_h, \sigma_{ag})$} constructing \mbox{$\A_{\E \land \E_h}$} as \mbox{$\A_{\E} \times \A_{\E_h}$} in polynomial time in $|\A_{\E_h}|$, and hence in $|h|$.

    By Theorem~\ref{thm:pass-and-dom} we can check that an agent using $\sigma_{ag}$ is actively responsible for $\omega$ under $\E$ by reducing to $\textsc{CheckWin}(\omega, \E, \sigma_{ag}) \land \textsc{ExistsWeak}(\lneg \omega, \E)$, giving \pspace and \twoexptime membership in $|\omega|$ and $|\E|$, respectively, and polynomial complexity in $|\sigma_{ag}|$. Hardness follows by reducing from checking if $\sigma_{ag}$ is winning for $\omega$ under $\E$, which is \pspace- and \twoexptime-hard in $|\omega|$ and $|\E|$, respectively, by Theorem~\ref{thm:check-win-complexity}. That is,  $\sigma_{ag}$ is winning for $\omega$ under $\E$ iff the agent is actively responsible for $\omega \land \lneg y$ under $\sigma_{ag}$ and $\E$, where $y$ is a new atom under agent's control. 
\end{proof}

\end{document}